\documentclass[lettersize,journal]{IEEEtran}
\usepackage{amsmath,amsfonts}
\usepackage{algorithmic}
\usepackage{algorithm}
\usepackage{array}
\usepackage{textcomp}
\usepackage{stfloats}
\usepackage{url}
\usepackage{verbatim}
\usepackage{graphicx}
\usepackage{cite}
\usepackage{graphicx}
\usepackage{amssymb}
\usepackage{booktabs}
\usepackage{xspace}

\newcommand{\figref}[1]{Figure~\ref{#1}}
\newcommand{\tabref}[1]{Table~\ref{#1}}
\newcommand{\secref}[1]{Section~\ref{#1}}

\makeatletter
\DeclareRobustCommand\onedot{\futurelet\@let@token\@onedot}
\def\@onedot{\ifx\@let@token.\else.\null\fi\xspace}

\def\eg{\emph{e.g}\onedot} 
\def\ie{\emph{i.e}\onedot}

\def\etal{\textit{et al}\onedot}
\makeatother

\usepackage[font=normalsize,labelfont=sf,textfont=sf]{subcaption}
\usepackage[inline]{enumitem}
\usepackage{multirow}
\usepackage{adjustbox}
\usepackage{xcolor}

\usepackage[capitalize]{cleveref}
\crefname{section}{Sec.}{Secs.}
\Crefname{section}{Section}{Sections}
\Crefname{table}{Table}{Tables}
\crefname{table}{Tab.}{Tabs.}

\begin{document}

\title{Mitigating Bias in Facial Analysis Systems by Incorporating Label Diversity}



\author{
        Camila Kolling, 
        Victor Araujo, 
        Adriano Veloso, 
        and Soraia Raupp Musse 
\thanks{
This study was financed in part by the Coordenação de Aperfeiçoamento de Pessoal de Nivel Superior – Brasil (CAPES) – Finance Code 001. \
textit{(Corresponding author: Camila Kolling.)}}
\thanks{Camila Kolling, Victor Araujo and Soraia Raupp Musse are with the Department of Computer Science, Pontifícia Universidade Católica do Rio Grande do Sul (PUCRS), Porto Alegre Av. Ipiranga 6681 Partenon, Brazil (email: camila.kolling@edu.pucrs.br; victor.araujo@edu.pucrs.br; soraia.musse@pucrs.br).}
\thanks{Adriano Veloso is with the Department of Computer Science, Universidade Federal de Minas Gerais (UFMG), Belo Horizonte Av. Antônio Carlos 6627 Pampulha, Brazil (email: adrianov@dcc.ufmg.br).
The authors would like to thank CNPq and CAPES for partially funding this work.
}
\thanks{Digital Object Identifier xx/TIP.xx.xxx}
}


\markboth{IEEE TRANSACTIONS ON IMAGE PROCESSING}%
{Shell \MakeLowercase{\textit{et al.}}: A Sample Article Using IEEEtran.cls for IEEE Journals}


\maketitle

\begin{abstract}
Facial analysis models are increasingly applied in real-world applications that have significant impact on peoples' lives.
However, as literature has shown, models that automatically classify facial attributes might exhibit algorithmic discrimination behavior with respect to protected groups, potentially posing negative impacts on individuals and society.
It is therefore critical to develop techniques that can mitigate unintended biases in facial classifiers.
Hence, in this work, we introduce a novel learning method that combines both subjective human-based labels and objective annotations based on mathematical definitions of facial traits.
Specifically, we generate new objective annotations from two large-scale human-annotated dataset, each capturing a different perspective of the analyzed facial trait.
We then propose an ensemble learning method, which combines individual models trained on different types of annotations.
We provide an in-depth analysis of the annotation procedure as well as the datasets distribution.
Moreover, we empirically demonstrate that, by incorporating label diversity, our method successfully mitigates unintended biases, while maintaining significant accuracy on the downstream tasks.
\end{abstract}


\section{Introduction}
\label{sec:intro}

\IEEEPARstart{I}{n}  recent years, artificial intelligence (AI) has been incorporated into a large number of real-world applications.
%
By integrating automated algorithm-based decision-making systems one might expect that the decisions will be more objective and fair.
Unfortunately, as several approaches have shown~\cite{chen2021understanding, denton2019image}, this is not always the case. 
Given that AI algorithms are trained with historical data, these prediction engines may inherently learn, preserve and even amplify biases~\cite{zhao2017men}. 
Special attention has been devoted to facial analysis applications since in the biometric modality performance differentials mostly fall across points of sensitivity (\eg race and gender)~\cite{drozdowski2020demographic}.

Particularly, the human face is a very important research topic as it transmits plenty of information to other humans, and thus possibly to computer systems~\cite{drozdowski2020demographic}, such as identity, emotional state, attractiveness, age, gender, and personality traits.
Additionally, the human face has also been of considerable interest to researchers due to the inherent and extraordinarily well-developed ability of humans to process, recognize, and extract information from others' faces~\cite{little2014facial}.
Faces dominate our daily situations since we are born, and our sensitivity  to them is strengthened every time we see a face under different conditions~\cite{little2014facial}.

With the vast adoption of those systems in high-impact domains, it is important to take fairness issues into consideration and ensure sensitive attributes will not be used for discriminative purposes.
In this work, we focus our attention on two main aspects of facial analysis, \textit{attractiveness} and \textit{facial expression}, both of which have a powerful impact on our lives.
For instance, the pursuit for beauty has encouraged trillion-dollar cosmetics, aesthetics, and fitness industries, each one promising a more attractive, youthful, and physically fitter version for each individual~\cite{cutler2021science}.
Beauty also seems to be an important aspect of human social interactions and matching behaviors, in which more attractive people appear to benefit from higher long-term socioeconomic status and are even perceived by peers as ``better'' people~\cite{cutler2021science}.

Simultaneously, facial expression is one of the most powerful, instinctive and universal signals for human beings to convey their emotional state and intentions~\cite{darwin2015expression}.
Often the face express an emotion before people can even understand or verbalize it.
Mehrabian \etal~\cite{mehrabian1974approach} shows that 55\% of messages regarding feelings and attitudes are conveyed through facial expression, 7\% of which are in the words that are spoken, and the rest of which are paralinguistic.

In part because of its importance and potential uses as well as its inherent challenges, automated attractiveness rating and facial expression recognition have been of keen interest in the computer vision and machine learning communities.
Several approaches have proposed methods for automatically assessing face attractiveness and expressions through computer analysis~\cite{ma2021robust, fasel2002head}.
However, as already shown in previous work~\cite{sattigeri2019fairness, ramaswamy2021fair, chen2021understanding}, models that are trained to automatically analyze facial traits
might exhibit algorithmic discrimination behavior with respect to protected groups, potentially posing negative impacts on individuals and society. 

Despite several advances towards understanding and mitigating the effect of bias in model prediction, limited research focused on
augmenting the label diversity of the annotations.
Thus, in this work, we introduce a novel learning method to mitigate such fairness issues.
We propose a method to generate and combine different types of label annotations, such as the subjective human-based labels, and the objective labels based on geometrical definitions of attractiveness and facial expression.
We hypothesize that introducing diversity into the decision-making model by adding mathematical and possibly unbiased notions in the label dimension should reduce the biases.
To the best of our knowledge, this is the first time a pre-processing debiasing method combines ``objective'' (mathematical) labels and ``subjective'' (human-based) annotations.

To summarize, our key contributions are as follows:
\begin{enumerate*}[label=(\arabic*)]
    \item We generate new annotations based on specific traits of the human face. 
    \item We propose a novel method that aims to mitigate biases by incorporating diversity into the training data for two tasks: attractiveness classification and facial expression recognition.
    However, we note that our approach extends to any tasks that has objective measures as well as subjective human labeling;
    \item We show that our approach is not dependent upon the data distribution of the novel annotations. 
    Moreover, we show that the models trained on mathematical notions achieve a better fairness metric
    compared to the models trained on only the human-based labels;
    \item Finally, using our method, we are able to achieve improvements in the fairness metrics over the baselines, while maintaining comparable accuracy.
\end{enumerate*}

\section{Related Work}
\label{sec:related_work}

We address the literature study according to three aspects: fairness in deep learning, attractiveness, and facial expression recognition (FER).


\subsection{Fairness in Deep learning}
\label{subsec:fairness_in_dl}

Machine learning algorithms can learn bias from a variety of different sources.
Everything from the data used to train it, to the people who are using this tool, and even seemingly unrelated factors can contribute to AI bias~\cite{mehrabi2019survey}.
Technical approaches to mitigate fairness issues may be applied to the training data (1), prior to modelling, known as \textit{pre-processing}; at the point of modelling (2), named as \textit{in-processing}; or at test time (3), after modelling, called \textit{post-processing}.

The pre-processing approaches argue that the issue is in the data itself, as the distributions of specific sensitive variables may be biased and/or imbalanced.
Thus, pre-processing approaches tend to alter the distribution of sensitive variables in the dataset itself. 
More generally, these approaches perform specific transformations on the data with the aim of removing discriminative attributes from the training data~\cite{celis2019improved}.
Our work fits into the \textit{pre-processing} approach since we add new ground-truth annotations for which different models are optimized.
In contrast with the pre-processing approaches, the in-processing ones argue that the fairness issue may be in the modelling technique~\cite{caton2020fairness}. 
Usually, these approaches tackle fairness issues by adding one or more fairness constraints into the model optimization functions towards maximizing performance and minimizing discriminative behavior.
Finally, the post-processing approaches recognize that the actual output of a model may be unfair to one or more protected variables~\cite{caton2020fairness}.
Thus, post-processing approaches tend to apply transformations to the model output to improve fairness metrics.

\subsection{Attractiveness}
\label{subsec:attractiveness}

Even though attractiveness has a considerable influence over our lives, which characteristics make a particular face attractive is imperfectly defined.
In modern days a common notion is that judgments of beauty are a matter of subjective opinion.
However, it has been shown that there is a very high agreement between groups of raters belonging to the same culture and even across cultures~\cite{cunningham1995their, langlois2000maxims}, and that people might share a common taste for facial attractiveness~\cite{kagian2007humanlike}.
Thus, if different people can agree on which faces are attractive and which faces are not attractive when judging faces of different ethnic backgrounds, then this indicates that people all around the globe use similar features or criteria when making up their judgments.

The earliest facial attractiveness predictors are based on traditional machine learning methods.
As well as for other computer vision tasks, deep learning brought great performance improvements for facial beauty assessment~\cite{gan2014deep, gray2010predicting}.
However, as recently demonstrated~\cite{ramaswamy2021fair, sattigeri2019fairness}, these approaches were shown to discriminate against certain groups.
Some pre-processing approaches were proposed to mitigate such behavior.
For instance, in the work of Sattigeri \etal~\cite{sattigeri2019fairness}, the authors use a generative model to create a dataset that is similar to a given dataset, but results in a model that is more fair with respect to protected attributes.
Another example is the work of Ramaswamy \etal~\cite{ramaswamy2021fair}, which modifies the generative model to create new instances by independently altering specific attributes (\eg removing glasses).
Both works expand the training dataset from 2 to 3 $\times$ its original size.

In this work, to extract the so-called \textit{objective} annotations of the attractiveness based on facial traits, we follow the work of Schmid \etal~\cite{schmid2008computation} and use three predictors that have been proposed in literature, and were empirically shown to correlate with human attractiveness ratings: Neoclassical Canons, Face Symmetry, and Golden Ratios.
Neoclassical canons were proposed by artists in the renaissance period as guides for drawing beautiful faces~\cite{farkas1985vertical}.
The basic idea behind this definition is that the proportion of an attractive face should follow some predefined ratios.
Facial symmetry has been shown as an important factor for attractiveness~\cite{rhodes2006evolutionary}.
It has many different definitions~\cite{schmid2008computation, rhodes2006evolutionary, kowner1996facial, perrett1999symmetry}, but it generally refers to the extent that one half of an image (\eg, face) is the same as the other half.
In our work, we follow the definition presented in the work of Schmid \etal~\cite{schmid2008computation}, which define the axis of symmetry to be located vertically at the middle of the face.
Finally, the golden ratio theory defines that faces that have features with ratios close to the golden ratio proportion are perceived as more attractive~\cite{gunes2011survey}.
The golden ratio is approximately the ratio of $1.618$~\cite{gunes2011survey}.
We refer to the work of Schmid \etal~\cite{schmid2008computation} for additional details on the methods used in this work.



\subsection{Facial Expression Recognition}
\label{subsec:fer}

Facial expression is one of the most powerful and instinctive means of communication for human beings.
In the past decade, much progress has been made to build computer systems to understand and use this natural form of human communication~\cite{grafsgaard2013automatically, fasel2002head, ma2021robust}.
Usually such systems are treated as a classification problem, and the basic set of emotions (classes) are defined as happiness, surprise, anger, sadness, fear, disgust and neutral (no emotion)~\cite{tian2001recognizing}.
As in other related fields, deep learning has improved the performance of FER systems, and some works~\cite{xu2020investigating, chen2021understanding} have recently focused on understanding and mitigating biases in such systems.
Li and Deng~\cite{li2020deep} observed that disgust, anger, fear, and surprise are usually underrepresented classes in datasets, thus being harder to learn compared to the majority classes.

Moreover, some works have shown slight differences in perception regarding some expressions in female and male faces.
For instance, women were shown to be generally seen as happier than men~\cite{steephen2018we}.
Becker \etal~\cite{becker2007confounded} demonstrated that people are faster and more accurate at detecting angry expressions on male faces and happy expressions on female faces.
Denton \etal~\cite{denton2019image} find that a smiling classifier 
is more likely to predict ``smiling'' when eliminating a person’s beard or applying makeup or lipstick to the image while keeping everything else unmodified.

In our work, to extract the \textit{objective} annotations of facial expressions, we adopt the Facial Action Coding System, which consists of facial action units (AUs)~\cite{ekman1978action} that objectively code the muscle actions typically seen for many facial expressions~\cite{du2014compound}.
Recently, in the FER field, the idea of using the relationship among multiple labels has been explored~\cite{chen2021understanding}.
In the context of objective labels to mitigate fairness issues, Chen and Joo~\cite{chen2021understanding} leads the initiative by proposing an in-processing approach that incorporates the triplet loss to embed the dependency between AUs and expression categories.
However, differently than previous work, we propose to use AUs in the pre-processing step.
Moreover, we show that our method not only successfully mitigates biases for FER, but also for rating attractiveness, and future work could expand it to other tasks that involve subjective labels.


\section{Proposed Method}
\label{sec:method}

Previous work~\cite{cunningham1995their} suggests that there is no single feature or dimension that determines attractiveness, and that attractiveness is the result of combining several features, which individually represent different aspects of a persons' face.
Moreover, this theory indicates that some facial qualities are perceived as universally (physically) attractive.
Similarly, facial expressions can be seen as a multi-signal system~\cite{revina2021survey}, and have been shown to posses universal meaning, regardless of culture and gender~\cite{ekman1993facial, ekman1976pictures}.
Based on these premises, we propose a method that combines several models trained on two main concepts: one based on different geometrical traits (\textit{objective annotations}) and another based on human judgment (\textit{subjective annotations}).

Our proposed method consists of three main steps:
(1) we generate annotations based on different mathematical notions.
(2) next, we train one machine learning model for each of the mathematical concepts, and one model with the original human-based annotations.
(3) finally, we aggregate all the models into an ensemble framework.
Our main hypothesis is that by combining the objective (geometrically-based), and the biased and subjective (human-based) notions we can effectively reduce the effect of discrimination on the system.
Our goal is to create a diverse set of decision-making algorithms that when combined can produce a fairer system.

To measure the discriminative behavior, we use different metrics according to the related work of the two tasks explored in this work.
For the attractiveness classification~\cite{sattigeri2019fairness, ramaswamy2021fair}, we use the metric of Equality of Opportunity ($\Delta EoO$), which is defined~\cite{sattigeri2019fairness} as the difference of conditional false negative rates across groups.
For the FER~\cite{chen2021understanding}, we use the Calders-Verwer discrimination score~\cite{calders2010three} ($\Delta Disc$), which is defined as the difference between conditional probabilities of advantageous decisions for non-protected and protected members.




\begin{figure*}[!t]
    \centering
    \begin{subfigure}[b]{0.3\textwidth}
        \includegraphics[width=\textwidth,height=2.5cm]{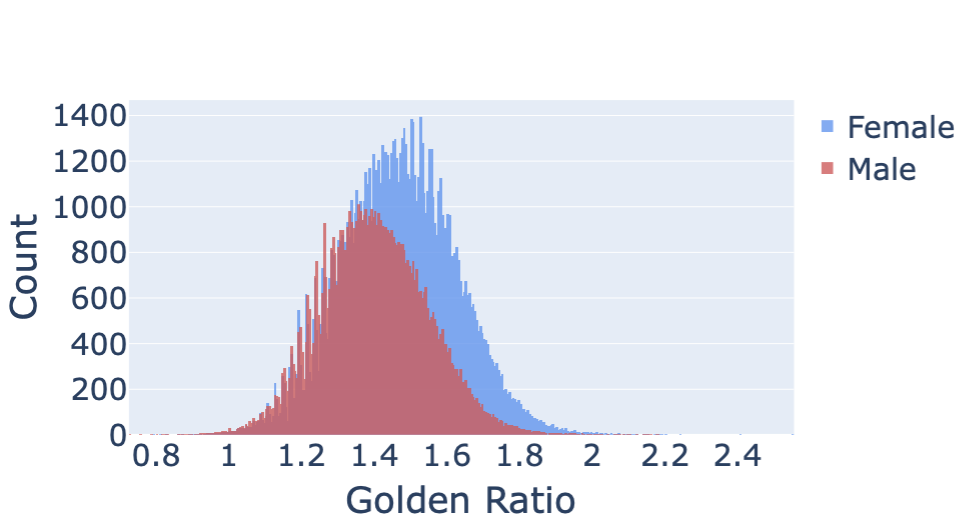}
        \caption{Golden ratio.}
        \label{fig:hist_gr}
    \end{subfigure}
    ~
    \begin{subfigure}[b]{0.3\textwidth}
        \includegraphics[width=\textwidth,height=2.5cm]{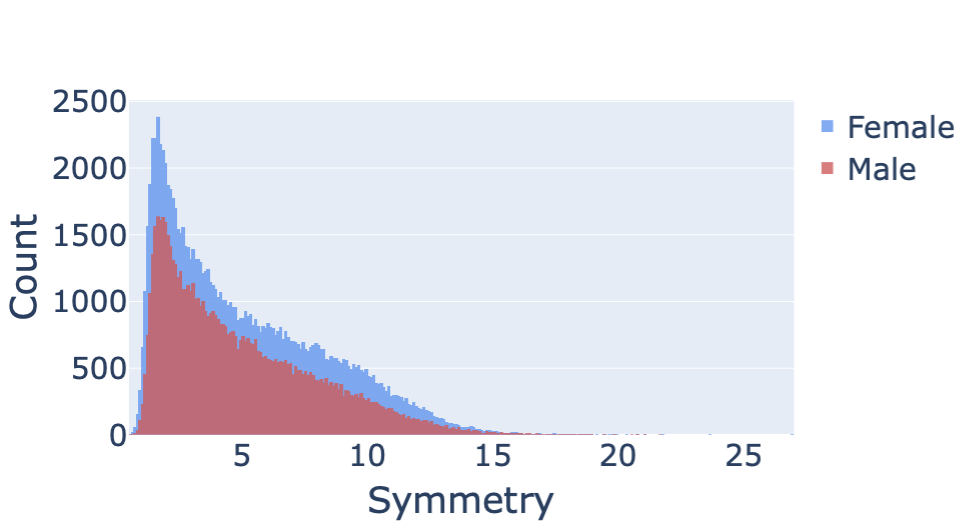}
        \caption{Symmetry.}
        \label{fig:hist_sym}
    \end{subfigure}
    ~
    \begin{subfigure}[b]{0.3\textwidth}
        \includegraphics[width=\textwidth,height=2.5cm]{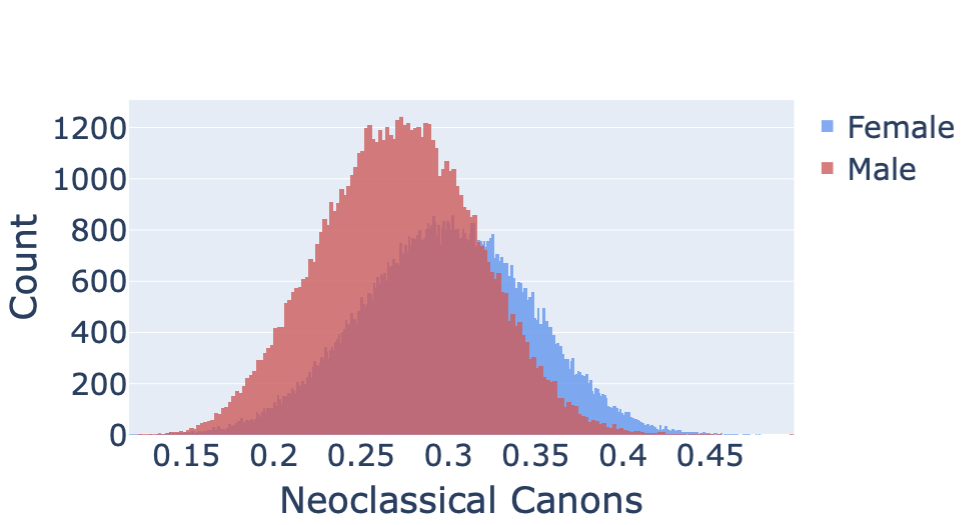}
        \caption{Neoclassical canons.}
        \label{fig:hist_neo}
    \end{subfigure}

    \caption{Histograms showing the data distributions per gender for the attractive attribute.}
    \label{fig:data_histograms}
\end{figure*}

\begin{figure*}
 \begin{subfigure}{0.24\textwidth}
     \includegraphics[width=\textwidth,height=3.3cm]{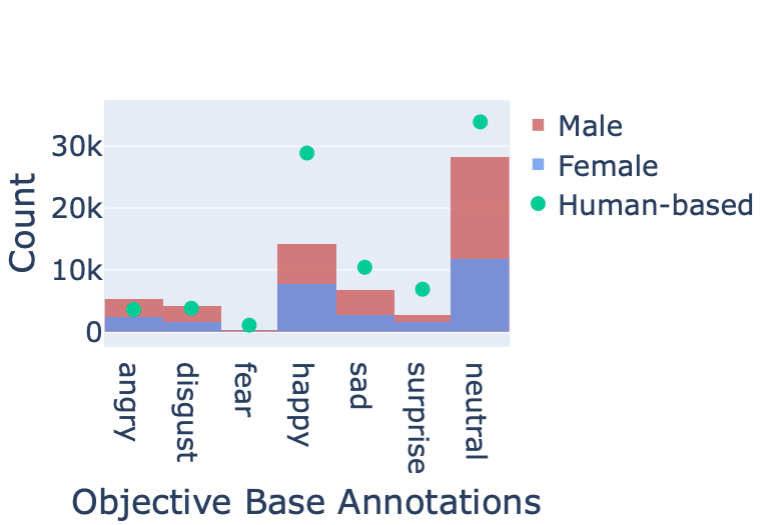}
     \caption{ObjBase.}
    \label{fig:hist_objbase}
 \end{subfigure}
 \hfill
 \begin{subfigure}{0.24\textwidth}
     \includegraphics[width=\textwidth,height=3.3cm]{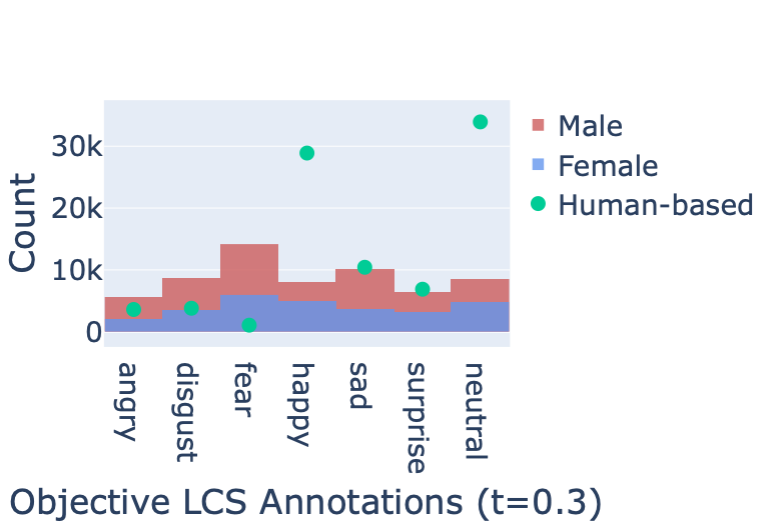}
     \caption{ObjLCS$_{0.3}$.}
    \label{fig:hist_objlcs03}
 \end{subfigure}
 \hfill
 \begin{subfigure}{0.24\textwidth}
     \includegraphics[width=\textwidth,height=3.3cm]{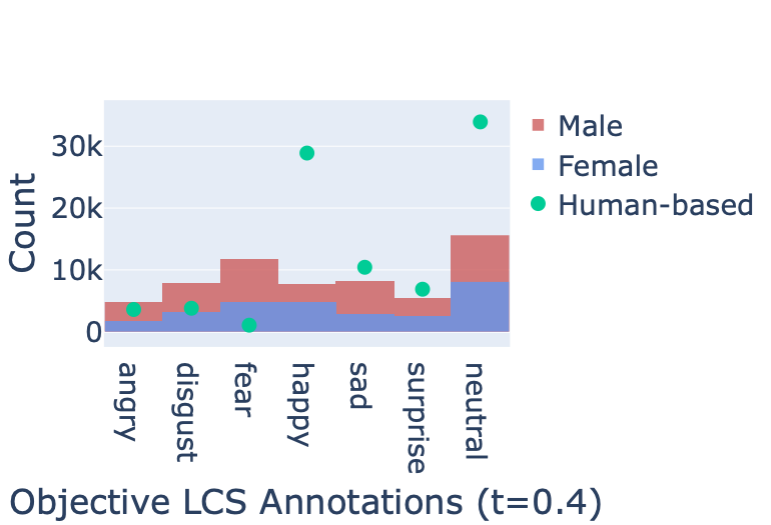}
     \caption{ObjLCS$_{0.4}$.}
    \label{fig:hist_objlcs04}
 \end{subfigure}
 \hfill
 \begin{subfigure}{0.24\textwidth}
     \includegraphics[width=1\textwidth,height=3.3cm]{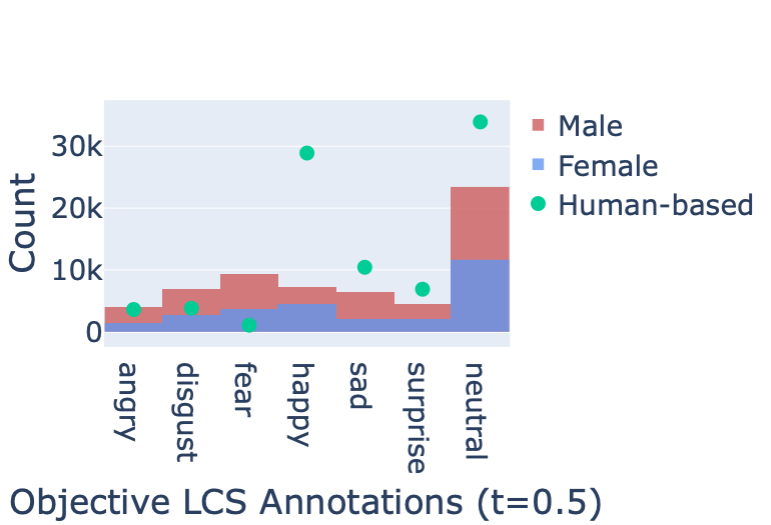}
    \caption{ObjLCS$_{0.5}$.}
    \label{fig:hist_objlcs05}
 \end{subfigure}

 \caption{Histograms showing the data distributions per gender for the facial expression attribute. 
Green dots represent the frequency of each facial expression for the human-based annotations. }
\label{fig:data_histograms_fer}
\end{figure*}

\subsection{Data Annotation}
\label{subsec:data_annotation}

The first step of our proposed method consists of generating annotations based on mathematical notions.
We next describe the process for each of the explored tasks.

\subsubsection{Attractiveness}
\label{subsubsec:data_annotation_attractiveness}

In this work, we use the attractive human-based labels from the CelebA dataset~\cite{liu2015faceattributes}, which
has been already studied by previous work~\cite{ramaswamy2021fair, sattigeri2019fairness} in the context of fairness.
We purposefully selected a dataset containing celebrities since previous work already claimed that some aspects of famous people might influence the way people rate attractiveness~\cite{thwaites2012impact}.
For instance, it is suggested by Thwaites \etal~\cite{thwaites2012impact} that the humor and personality associated with a specific character make the celebrity attractive.
Additionally, previous work~\cite{sattigeri2019fairness, ramaswamy2021fair} already studied and demonstrated fairness issues regarding the attractive feature in this dataset.

Our first step is composed of extracting facial landmarks from the images.
Given that the attractiveness measures we are using in this step are based on geometrical traits and landmarks of the face, and in order to avoid miscalculation, we discard some lateral facial poses.
Specifically, we remove images in which the difference between the euclidean distance of the eyes
is less than a given threshold.
We empirically tested several thresholds on thousand images and found that the best one was $\beta = 10$.
At the end of this process of discarding lateral images from the original CelebA dataset, we kept $137,048$ of the $162,770$ training images ($84.19\%$), and $16,896$ of the $19,867$ validation images ($85.04\%$).

Once we are able to detect all the necessary facial landmarks in mostly frontal images, we calculate the mathematical notions of attractiveness for all the remaining images.
\figref{fig:data_histograms} shows the dataset distribution for female and male for each calculated metric.
Even though the distributions for the same metric are similar across gender, the high peaks across the metrics are slightly different.
Additionally, each metric has its own ideal (target) value. 
While golden ratio defines the best ratio as $1.618$, symmetry and neoclassical canons define it as $0$.
We note that neoclassical canons is the curve which posses the most distant peak from the target value.
We hypothesize that this attractiveness metric may be too restrictive, thus not perfectly reflecting the real distribution of the data.
This might be related to the fact that neoclassical canons is based on artistic concepts, as mentioned by Farkas \etal~\cite{farkas1985vertical}.

Since all of the attractiveness definitions generate a unique continuous value describing the attractiveness of each person, and the CelebA dataset has binary attribute annotations, to obtain consistency we define five different ranges for each mathematical attractiveness measure.
These ranges correspond to the amount of variation each metric tolerates.
For golden ratio, since its ideal value is $1.618$, we define a delta ($\delta$) value that defines the range for which one is considered attractive, \eg, for a hypothetical $\delta=0.1$, we consider all images that possess golden ratio from $1.518$ ($1.618 - \delta$) to $1.718$ ($1.618 + \delta$) as attractive.
In contrast, since symmetry and neoclassical canons establish the ideal value as $0$, and negative values for both metrics are infeasible, we use a threshold $t$ that defines the range of attractive people from $0$ to $t$, \eg for $t=4$ a person is considered attractive if it contains a proportion below $4$.

Therefore, the higher the $\delta$ or the $t$, the more people will fit into the attractive category.
Our goal when choosing $\delta$ and $t$ is to obtain the most balanced attractive distribution possible.
Thus, we generate, for each metric, five new binary ground-truth labels for attractiveness, one for each chosen range ($\delta$ or $t$).
It is important to note that we use the annotations provided by the CelebA for the sensitive attribute.
Furthermore, the subjective annotations~\cite{bohlen2017server} are the ones we obtain when using the original attractiveness labels from the CelebA. 
We solely generate annotations for the objective definition of attractiveness based on geometrical traits of the human face.

\subsubsection{Facial Expression Recognition}
\label{subsubsec:data_annotation_fer}

For the FER system, we use all the pre-processing procedures and datasets provided in the work of Chen and Joo~\cite{chen2021understanding}.
We highlight the fact that, in their work, the training datasets and the test splits differ.
Thus, ExpW~\cite{zhang2015learning, zhang2018facial} was used for training, and CFD~\cite{ma2015chicago} was used as test set.
Moreover, all models are treated as a binary classification, \ie, happy/unhappy, and the unhappy class is defined as all the instances that are not annotated as happy in the original human-based labels.

For generating the objective annotations, our first step is composed of extracting the AUs for all images in the training dataset.
We extract both the presence (as a binary feature) as well as the intensity (as a float feature) of each AU.
Next, we create two algorithms, one which we named `ObjBase', based mainly on the detection of AUs, and a second named `ObjLCS', which is also based on intensities.
For the first one, we simply test whether the combination of AUs for a specific facial expression exists, \eg whether all the AUs that compose an expression~\cite{ekman1978action} are active for a particular image.
Thus, we only annotate the image as containing a particular facial expression if all the AUs that compose that expression are activated.
In case two facial expressions are detected, which happens in 12\% of images, we average the intensity of the AUs that compose the facial expressions, and annotate the one which contains the highest average.

However, requiring that all AUs be active is a very hard constraint for some expressions, such as `fear' which is composed of a combination of seven AUs.
This can be observed in~\figref{fig:hist_objbase} since almost no instance (\ie, 0.3\%) is labeled as containing the `fear' expression.
Hence, next, we propose a second algorithm for annotating the expressions in an objective way.
We use the Longest Common Subsequence (LCS)~\cite{velusamy2011method} method, which is a function whose purpose is to count the number of operations required to transform one string (\ie, activated AUs) into another (\ie, AUs that compose a facial expression).
Our goal is to find the expression which contains the AUs closer to the detected ones,
not necessarily requiring that all AUs pertaining to that expression be active.

Using the LCS, we compare the AUs detected in the image with the AUs that represent each of the six basic emotions as found by previous literature~\cite{mavadati2013disfa}.
However, using this method might select more than one expression for each image.
In this case, we follow the work of Peres \etal~\cite{peres2021towards} which calculates the euclidean distance between the intensity of the AUs detected by each facial expression selected and the ones defined by Ekmann \etal~\cite{ekman1978action}.
The expression that results in the lowest Euclidean distance is the annotated one.

Nevertheless, extracting the neutral expression individually using the LCS algorithm is not possible since we try to approximate the expression which possesses the AUs closer to the active AUs, while the neutral expression implies absolutely no AU is active.
To solve this, we use a threshold $t$ that determines a neutral expression if the average intensity of the AUs of the detected facial expression is less than $t$.
We define three possible thresholds $t$ ($0.3$, $0.4$ and $0.5$), with distributions depicted on Figures~\ref{fig:hist_objlcs03}, \ref{fig:hist_objlcs04} and \ref{fig:hist_objlcs05}, respectively.
The higher the $t$, the more concentrated the distribution is to the neutral expression (\ie, less evenly distributed among the six facial expressions).
The green dots represent the distribution of the expressions for the human-based annotations ($H$).
We note that all thresholds $t$ provide a more distributed labels per facial expression than the one annotated by humans, \ie, for the human-based labels 70.9\% of the instances are considered as happy or neutral.
Thus, all expressions are more well-represented and balanced in the objective annotations.

\subsection{Model Training}
\label{subsec:train_models}

In our experiments for the attractive attribute, we use ResNet-$18$~\cite{he2016deep} as the base architecture, while for the FER, following previous work~\cite{chen2021understanding}, we use ResNet-$50$ pre-trained on ImageNet~\cite{russakovsky2015imagenet}.
The inputs of the ResNet-$18$ model are $128 \times 128$ colored images, and $224 \times 224$ for the ResNet-$50$.
All models were trained with the cross entropy loss and Adam~\cite{kingma2014adam} optimizer.
The learning rate (LR) was set to $1e-3$ for the attractiveness classification, and $1e-4$ for the FER.
We use LR scheduler for the former one, which reduces the initial value by $0.1$ when the validation loss does not improve for $10$ epochs.
For the latter one, we reduce the LR by $0.1$ every $6$ epochs for the initial LR.

\subsection{Ensemble}
\label{subsec:ensemble}

The last main step in our proposed method is combining models trained on different perspectives.
This step has two main motivations:
\begin{enumerate*}[label=(\arabic*)]
    \item~most recent approaches replace several human decision-makers with a single algorithm, such as COMPAS for recidivism risk estimation~\cite{angwin2016there}. However, in high-stake real-world applications, the decision is taken from multiple human beings. Thus, we argue that one could introduce diversity into machine decision making by instead training a collection of algorithms, each capturing a different perspective about the problem solution, and then combining their decisions in some ensemble manner (\eg, simple or weighted majority voting);
    \item~the rich literature on ensemble learning, where a combination of a diverse ensemble of predictors have been shown (both theoretically and empirically) to outperform single predictors on a variety of tasks~\cite{brown2005diversity}.
\end{enumerate*}
In this work, we implement bagging
using the following weighted process for each instance $x$ of the test set:

\begin{equation}
    f(x) = \sum_{i=1}^{M} \alpha_i \cdot o_{i}(x),
    \label{eq:bagging}
\end{equation}

\noindent
where $f(x)$ is the ensemble prediction for the instance $x$, $M$ is the number of individual models we combine, $o_{i}$ is the output of the $i_{th}$ model for instance $x$, and $\alpha_i$ represents the weight that the $i_{th}$ model has in the final ensemble output of $f(x)$.
Hence, each model has an influence in the final decision.

\section{Results}
\label{sec:results}

In this section we describe our main results for both tasks.
Each follow the same presentation of sections:
\begin{enumerate*}[label=(\arabic*)]
    \item we first report the distribution of the generated datasets;
    \item  we then show the result of the individual models on the test sets;
    \item next, we show the results when combining individual models
    into an ensemble;
    \item finally we compare our results with previous literature.
\end{enumerate*}

\begin{table}[!t]
    \scriptsize
    \begin{adjustbox}{width=\columnwidth}
    \small
    \centering
        \begin{tabular*}{1.1\columnwidth}{@{\extracolsep{\fill}}cc|ccc|ccc}
            \toprule
            Def. & $\delta$ or $t$ & $Y=1$ & M & F & $Y=0$ & M & F \\
            \midrule
            \multirow{5}{*}{$GR$}
            & 0.17 & 46\% & 32.5\% & 67.5\% & 54\% & 48.9\% & 51.1\% \\
             & 0.18 & 48\% & 32.9\% & 67.1\% & 52\% & 49.3\% & 50.7\% \\
             & 0.19 & 51\% & 33.4\% & 66.6\% & 49\% & 49.7\% & 50.3\% \\
             & 0.20 & 54\% & 33.8\% & 66.2\% & 46\% & 50.2\% & 49.8\% \\
             & 0.21 & 56\% & 34.2\% & 65.8\% & 44\% & 50.6\% & 49.4\% \\
            \midrule
            \multirow{5}{*}{$S$} & 4.0 & 47\% & 41.9\% & 58.1\% & 53\% & 40.9\% & 59.1\% \\
             & 4.2 &50\% & 42.0\% & 58.0\% & 50\% & 40.8\% & 59.2\%\\
             & 4.4 & 52\% & 42.1\% & 57.9\% & 48\% & 40.6\% & 59.4\% \\
             & 4.6 & 54\% & 42.2\% & 57.8\% & 46\% & 40.4\% & 59.6\% \\
             & 4.8 & 56\% & 42.3\% & 57.8\% & 44\% & 40.3\% & 59.7\% \\
            \midrule
            \multirow{5}{*}{$NC$}
            & 0.26 & 29\% & 56.5\% & 43.5\% & 71\% & 35.2\% & 64.8\% \\
             & 0.27 & 36\% & 54.8\% & 45.2\% & 64\% & 33.7\% & 66.3\% \\
             & 0.28 & 44\% & 53.2\% & 46.8\% & 56\% & 32.1\% & 67.9\% \\
             & 0.29 & 52\% & 51.7\% & 48.3\% & 48\% & 30.1\% & 69.9\% \\
             & 0.30 & 60\% & 49.8\% & 50.2\% & 40\% & 28.2\% & 71.8\% \\
             \midrule
             $H$ & - & 53\% & 23.3\% & 76.7\% & 47\% & 61.7\% & 38.3\% \\
            \bottomrule
        \end{tabular*}
    \end{adjustbox}
    \caption{Dataset distribution for each attractiveness notion per range $\delta$ or threshold $t$.
    For both attractive ($Y=1$) and not attractive ($Y=0$), we also add their distribution with respect to gender, \ie, male (M) and female (F).
    $GR$, $S$, $NC$ and $H$ correspond to the different attractiveness definitions (Def.): golden ratio, symmetry, neoclassical canons, and human perception (\ie, the ones obtained from the dataset itself), respectively.
    We highlight that the distribution is similar across train, validation and test sets.}
    \label{tab:data_ann_distr}
\end{table}

\subsection{Attractiveness}
\label{subsubsec:results_attr}

\subsubsection{Dataset Distribution}
\label{subsubsec:dataset_distribution_attr}

Our objective when generating different $\delta$ and $t$ choices is to analyze whether slightly altering the dataset distribution heavily affects the models' behavior.
In~\tabref{tab:data_ann_distr} we show the distribution of the dataset regarding the new attractiveness measures for each attractiveness range $\delta$ or threshold $t$, also scattered across the sensitive attribute of gender expression.
We also added the new distribution of the CelebA dataset ($H$) when removing lateral facial poses.
We first observe from~\tabref{tab:data_ann_distr} that the target attribute has a distribution close to $50\%$ for at least one option of $\delta$ and $t$ for all attractiveness metrics.
For instance, the range $\delta=0.19$ for golden ratio has $51\%$ attractive people, the thresholds $t=4.2$, for symmetry, $t=0.29$ for neoclassical canons, have $50\%$  and $52\%$ attractive people, respectively.
We also note that the gender attribute varies according to the target attribute and threshold, \ie, when the target attribute is close to $50\%$, the distribution of male/female is also close to $50\%$.
The only exception are the human-based ($H$) labels.

\subsubsection{Individual Models}
\label{subsec:result_indiv_models_attr}

In this section, we describe the results when individually training the models.
Specifically, we train one model for each range $\delta$ or threshold $t$ and attractiveness definition as depicted in~\tabref{tab:data_ann_distr}.
We then evaluate the models on the human-based attractiveness concept, using CelebA original test set annotations.
For reproducibility, we run each model over three seeds, and report the average performance.
Our goal is to verify whether testing ``objective'' models on the CelebA test set actually reduces the fairness metric compared to the model originally trained on its original and ``subjective'' annotations.
\tabref{tab:all_results_individual} shows the results for models trained on golden ratio, symmetry, and neocanons, respectively, evaluated on CelebA test set for the sensitive attribute \textit{male}.
We also added a row named $H$, which depicts the average result of the models trained and evaluated on CelebA annotations.
We highlight the fact that all the models were trained only using the (mostly) frontal images, \ie, all models were trained on the same set of images, however, each one used a different ground-truth annotation during training.

We first note a trade-off between accuracy (`Overall' Accuracy column) and fairness ($\Delta EoO$ column), as previously discussed in the literature~\cite{haas2019price}.
This is especially the case for models trained on objective annotations, compared to the one trained on CelebA, which obtained a better fairness result (lower $\Delta EoO$) but close to random overall accuracy ($50\%$ in a binary classification problem).
However, the low accuracy is expected since they were not trained to capture subjective human-like patterns, instead, they were trained to detect mathematical definitions of attractiveness.
Furthermore, we notice that the $\Delta EoO$ is much lower on the models based on geometrical traits than the one trained on human-based labels.
In other words, \textit{all} the models trained on geometrical concepts of attractiveness are much less discriminative in the chosen fairness metric than the one trained on subjective notions, regardless of the choice of threshold $\delta$ or $t$.

Therefore, we can conclude that individually training models on the geometric notions of attractiveness improve the fairness metrics for all test set evaluations.
This supports our claim that models trained to perceive mathematical notions of attractiveness in fact mitigate some forms of biases.
However, even though our goal is to add the fairness constraint to the unfair decision-making process, we do not wish to reduce the accuracy to a random-choice level since this results in a useless model that would be misclassifying half the instances.
Thus, we next combine all models into an ensemble.

\begin{table}[!t]
    \begin{adjustbox}{width=1\columnwidth}
    \small
    \centering
        \begin{tabular*}{1.3\columnwidth}{@{\extracolsep{\fill}}c|c|ccc|cc|c}
            \toprule
            & & \multicolumn{3}{c}{Accuracy} & & & \\
            Def. & $\delta$ or $t$ & Overall & $s=1$ & $s=0$& $\Delta$TPR & $\Delta$FPR & $\Delta$EoO \\
            \midrule
            \multirow{5}{*}{$GR$}
            & 0.17 & 0.552 & 0.514 & 0.613 & 0.155 & 0.247 & \underline{0.162}\\
            & 0.18 & 0.557 & 0.531 & 0.600 & 0.136 & 0.243 & \textbf{0.144} \\
            & 0.19 & 0.555 & 0.530 & 0.595 & 0.149 & 0.255 & 0.160\\
            & 0.20 & 0.556 & 0.541 & 0.580 & 0.136 & 0.240 & 0.150\\
            & 0.21 & 0.554 & 0.551 & 0.558 & 0.144 & 0.233 & 0.157 \\
            \midrule
            \multirow{5}{*}{$S$}
            & 4.0 & 0.510 & 0.495 & 0.534 & 0.081 & 0.008 & \textbf{0.082} \\
            & 4.2 & 0.510 & 0.501 & 0.525 & 0.083 & 0.014 & 0.085 \\
            & 4.4 & 0.509 & 0.504 & 0.515 & 0.090 & 0.018 & 0.091 \\
            & 4.6 & 0.507 & 0.509 & 0.503 & 0.089 & 0.019 & 0.090 \\
            & 4.8 & 0.509 & 0.518 & 0.494 & 0.092 & 0.014 & \underline{0.093} \\
            \midrule
            \multirow{5}{*}{$NC$}
            & 0.26 & 0.438 & 0.376 & 0.537 & 0.145 & 0.147 & \textbf{0.131} \\
            & 0.27 & 0.436 & 0.411 & 0.476 & 0.158 & 0.168 & 0.144 \\
            & 0.28 & 0.426 & 0.416 & 0.441 & 0.184 & 0.196 & \underline{0.167} \\
            & 0.29 & 0.434 & 0.456 & 0.397 & 0.175 & 0.186 & 0.158 \\
            & 0.30 & 0.442 & 0.501 & 0.348 & 0.179 & 0.204 & 0.166 \\
            \midrule
            $H$ & - & 0.807 & 0.800 & 0.820 & 0.176 & 0.275 & 0.292 \\
            \bottomrule
        \end{tabular*}
    \end{adjustbox}
    \caption{
    Results of models trained on different attractiveness notions and evaluated on CelebA test set.
    We show the average results across three different seeds for the sensitive attribute \textit{male}. 
    $GR$, $S$, $NC$ and $H$ corresponds to the different attractiveness definitions (Def.): golden ratio, symmetry, neoclassical canons, and human perception (\textit{i.e.}, the ones obtained from the dataset itself), respectively.
    We highlighted the best and underlined the worst average $\Delta$EO results.}
    \label{tab:all_results_individual}
\end{table}

\subsubsection{Ensemble Model}
\label{subsubsec:result_ensemble_attr}

\begin{figure*}[!t]
    \centering
    \begin{subfigure}[b]{0.45\textwidth}
        \centering
        \includegraphics[width=1\textwidth,height=4cm]{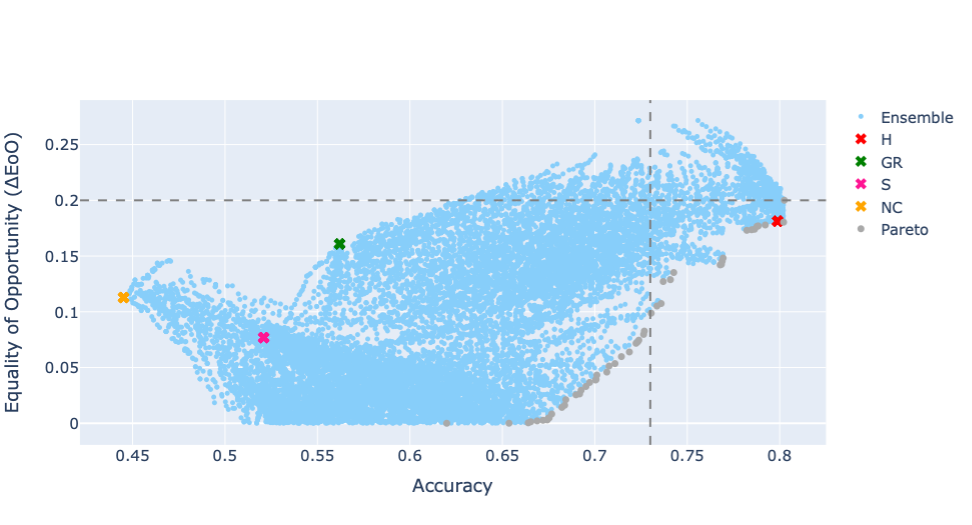}
        \caption{Best performing models.}
        \label{fig:ensemble_best_male_pareto}
    \end{subfigure}
    ~
    \begin{subfigure}[b]{0.45\textwidth}
        \centering
        \includegraphics[width=1\textwidth,height=4cm]{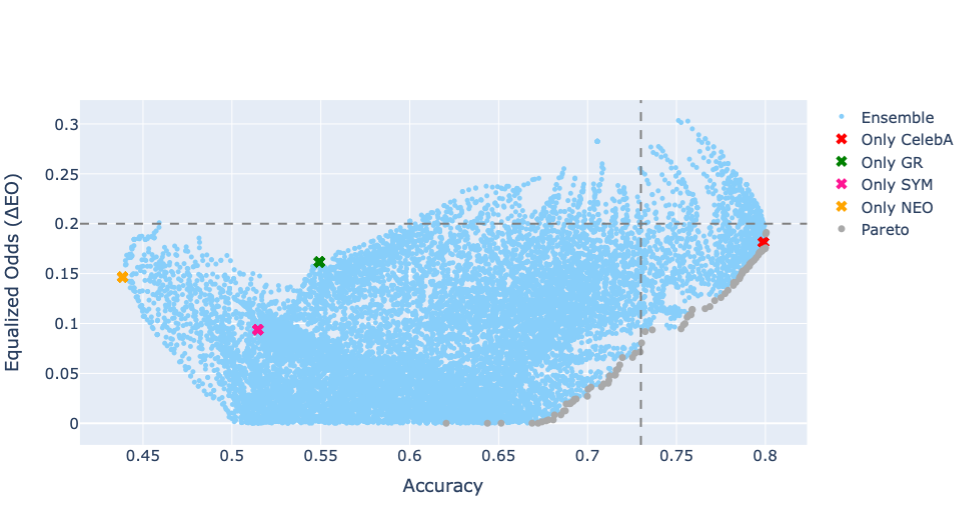}
        \caption{Worst performing models.}
        \label{fig:ensemble_worst_male_pareto}
    \end{subfigure}
    \caption{Results of different weighting values to compose the final ensemble for the attractive attribute. Plots on the first and second column correspond to the best and worst individual models with respect to $\Delta EoO$ in~\tabref{tab:all_results_individual}, respectively. We show the result of each ensemble with respect to accuracy ($x$ axis) and $\Delta EoO$ ($y$ axis) for the sensitive attribute \textit{male}. Light blue dots represent the \textit{ensemble} models, \ie, combining different definitions of attractiveness; red, green, pink and orange crosses indicate the model trained with only CelebA ($H$), golden ratio ($GR$), symmetry ($S$) and neoclassical canons ($NC$) annotations, respectively; finally, the gray dots represent the Pareto analysis.}
    \label{fig:results_ensemble_male}
\end{figure*}

The ensemble in this section combines four models, each trained on a different definition of attractiveness.
As previously described in~\secref{subsec:ensemble}, we used a weighted combination of the models, \ie, each individual model possesses a different influence over the final prediction.
\figref{fig:results_ensemble_male} shows the result of several weighting values per model for the sensitive attribute \textit{male}.
The plot on the left (\figref{fig:ensemble_best_male_pareto}) corresponds to the ensemble results for the selected best individual models, \ie, models which obtained best results with respect to $\Delta EoO$ in the CelebA test set, while the one on the right (\figref{fig:ensemble_worst_male_pareto}) depicts the result for the worst models.
The previous results showed an average result across three different runs, each with a different seed.
However, when combining the models in the ensemble we randomly chose one seed for all models.

We show the result of each ensemble with respect to accuracy ($x$ axis) and $\Delta EoO$ ($y$ axis) in~\figref{fig:results_ensemble_male}.
Each blue dot illustrates the result of one ensemble model (one weighted combination) of the four models.
We varied the weight of each individual (base) model from $0$ to $1$ with steps $0.1$.
Thus, at the end, we obtain more than $10,000$ possible combinations.
To best understand the results over the baseline, we also plotted the result of models trained with a single attractiveness definition.
Thus, the model trained only on human-based CelebA annotations is shown in red, and the ones trained only with mathematical concepts of attractiveness, such as golden ratio, symmetry and neoclassical canons, are shown in green, pink, and orange, respectively.
Finally, the gray dots represent the Pareto analysis, which is based on Pareto efficiency~\cite{iancu2014pareto}.
Pareto-optimal solution in multi-objective optimization delivers optimized performance across different objectives~\cite{iancu2014pareto}.
In this work, we wish to optimize for both accuracy and fairness.
Thus, the optimal solutions when maximizing accuracy and minimizing $\Delta EoO$ are the ones shown in gray.

We first observe that both~\figref{fig:ensemble_best_male_pareto} and~\figref{fig:ensemble_worst_male_pareto} show comparable curve results.
This suggests that individually combining the best and worst models into an ensemble have approximately the same result regarding overall accuracy and $\Delta EoO$.
Moreover, it reinforces our previous finding, in which the choice of $\delta$ and $t$ does not have a huge impact on the final result.
In contrast, as shown in both plots, the final result is heavily dependent on the weights each base model has in the final ensemble.
This can be directly inferred from how scattered the ensemble models are (light blue dots).
Therefore, from \figref{fig:results_ensemble_male} we can see that there is a wide range of possible models.
For instance, from the horizontal gray line, fixed at $\Delta EoO=0.2$~\cite{ramaswamy2021fair}, it is possible to obtain an overall accuracy from $\approx 0.62$ to more than $0.8$.
Simultaneously, it is also possible to obtain an ensemble whose accuracy is $0.73$~\cite{sattigeri2019fairness}, represented by the vertical gray line, whose $\Delta EoO$ varies from $0.1$ to less than $\approx 0.2$.
The decision upon which ensemble to choose from will depend heavily on the downstream application.

\begin{table}[!t]
    \begin{adjustbox}{width=1\columnwidth}
    \small
    \centering
        \begin{tabular*}{1.1\columnwidth}{@{\extracolsep{\fill}}l|ccc|c}
            \toprule
            & \multicolumn{3}{c}{Error Rate} & \multirow{2}{*}{$\Delta EoO$} \\
            & $s=0$ & $s=1$ & overall & \\
            \midrule
            Fairness GAN~\cite{sattigeri2019fairness} & 0.29 & 0.24 & 0.27 & 0.23 \\
            LSD~\cite{ramaswamy2021fair} & 0.21 & \textbf{0.18} & - & 0.20 \\
            \midrule
            $Ours_1$ & 0.30 & 0.27 & 0.29 & \textbf{0.05} \\
            $Ours_2$ & 0.23 & 0.22 & \textbf{0.22} & \textbf{0.14} \\
            $Ours_3$ & \textbf{0.20} & 0.20 & \textbf{0.20} & \textbf{0.17} \\
            \bottomrule
        \end{tabular*}
    \end{adjustbox}
    \caption{Comparison of Fairness GAN~\cite{sattigeri2019fairness}, Latent Space De-biasing (LSD)~\cite{ramaswamy2021fair}, and a set of the ensemble models we obtained from combining individual models trained on different attractiveness definitions, as shown in \figref{fig:results_ensemble_male}.}
    \label{tab:comparison_fairness_gan}
\end{table}

\subsubsection{Comparison with Prior Work}
\label{sunsubsec:comparison_sota_attr}

Finally, in this section, we compare our method with previous approaches.
In order to choose some ensembles over all possible combinations, we opted for selecting models included in the Pareto boundary for both plots.
Specifically, we sort the models with respect to fairness, and select the top $3$ models contained in both boundaries, \ie, first three models in the intersection of both Pareto boundaries.
We compare our method with previous debiasing approaches for the attractive attribute~\cite{sattigeri2019fairness, ramaswamy2021fair}.
\tabref{tab:comparison_fairness_gan} depicts the results.
We first note that all of our approaches have the lowest $\Delta EoO$, while maintaining significant accuracy compared to previous work.
Moreover, we show that all of our metrics are comparable or better than both Fairness GAN and LSD approaches, both of which incorporate $2$ to $3 \times$ more synthetic images to the original dataset.
Besides substantially increasing the computational complexity, these approaches may add low-quality and even unrealistic data.
Thus, we obtained a better trade-off between a given fairness metric ($\Delta EoO$) and accuracy compared to other pre-processing approaches for the attractive attribute.

\begin{table}[!t]
    \scriptsize
    \begin{adjustbox}{width=\columnwidth}
    \small
    \centering
        \begin{tabular*}{1.2\columnwidth}{@{\extracolsep{\fill}}cc|ccc|ccc}
            \toprule
            Data & Def. & $Y=1$ & M & F & $Y=0$ & M & F \\
            \midrule
            \multirow{5}{*}{ExpW}
            & ObjBase & 25.4\% & 64.8\% & 35.2\% & 74.6\% & 70.5\% & 29.5\% \\
            & ObjLCS$_{0.3}$ & 15.0\% & 62.0\% & 38.0\% & 85.0\% & 70.3\% & 29.7\% \\
            & ObjLCS$_{0.4}$ & 14.5\% & 61.9\% & 38.1\% & 85.5\% & 70.2\% & 29.8\% \\
            & ObjLCS$_{0.5}$ & 13.5\% & 62.1\% & 37.9\% & 86.5\% & 70.1\% & 29.9\% \\
            & $H$ & 33.1\% & 63.2\% & 36.8\% & 66.9\% & 71.9\% & 28.1\% \\
            \bottomrule
            CFD-Hap & $H$ & 36.3\% & 50.0\% & 50.0\% & 63.7\% & 50.0\% & 50.0\% \\
        \end{tabular*}
    \end{adjustbox}
    \caption{
    Dataset distribution for the happiness expression.
    We use ExpW~\cite{zhang2015learning, zhang2018facial} as training dataset, however, following Chen and Joo~\cite{chen2021understanding}, during training, we sample $20k$ instances and average the results over $5$ runs~\cite{chen2021understanding}.
    For both happy ($Y=1$) and unhappy ($Y=0$) expressions, we also add their distribution with respect to gender, \ie, male (M) and female (F).
    ObjBase and ObjLCS$_t$ correspond to the different objective definitions (Def.) for annotating the facial expressions using AUs, respectively, and $H$ represent the human perception.
    CFD-Hap corresponds to the test set used to evaluate models~\cite{chen2021understanding}.
    }
    \label{tab:fer_data_ann_distr}
\end{table}

\subsection{Facial Expression Recognition}
\label{subsec:results_fer}

\subsubsection{Dataset Distribution}
\label{subsubsec:dataset_distribution_fer}

\tabref{tab:fer_data_ann_distr} shows the distribution of the training and test dataset for the FER system.
We also added the distribution of the ExpW dataset ($H$).
We first note that, due to the the fact that we are dealing with a binary classification~\cite{chen2021understanding}, all expressions that are not annotated as happy, are considered as unhappy.
Thus, we end up with an imbalanced training dataset for both subjective and objective annotations.
The distribution of the test set was purposely modified~\cite{chen2021understanding} such as the allocation of happy and unhappy images between male and female would be the same.
We can visualize that the base algorithm for generating the `objective' labels is the closest one to the distribution of the the labels provided by humans.
Additionally, since the algorithm ObjLCS$_t$ generates labels that are more spread across all the six facial expressions, it ends up providing an even more imbalanced dataset with respect to the binary happiness attribute.

\begin{table}[!t]
    \begin{adjustbox}{width=1\columnwidth}
    \small
    \centering
        \begin{tabular*}{1\columnwidth}{@{\extracolsep{\fill}}l|ccc|c}
            \toprule
            & \multicolumn{3}{c}{Accuracy} & \\
            Def. & Overall & M & F & $\Delta Disc$ \\
            \midrule
            ObjBase & 0.926 $\pm$ 0.007 & 0.932 & 0.921 & 0.052 $\pm$ 0.019 \\
            ObjLCS$0.3$ & 0.826 $\pm$ 0.027 & 0.830 & 0.821 & 0.009 $\pm$ 0.021 \\
            ObjLCS$0.4$ & 0.829 $\pm$ 0.013 & 0.833 & 0.825 & 0.013 $\pm$ 0.022 \\
            ObjLCS$0.5$ & 0.816 $\pm$ 0.017 & 0.822 & 0.810 & 0.013 $\pm$ 0.027 \\
            \midrule
            $H$ & 0.935 $\pm$ 0.009 & 0.936 & 0.934 & 0.046 $\pm$ 0.025 \\
            \bottomrule
        \end{tabular*}
    \end{adjustbox}
    \caption{
    Results of models trained on different facial expression notions and evaluated on CFD test set.
    We show the average results across five different seeds for the sensitive attribute \textit{gender}~\cite{chen2021understanding}. 
    ObjBase and ObjLCS$t$ correspond to the different objective definitions (Def.) for annotating the facial expressions using AUs, respectively, and $H$ represent the human perception.
    We highlighted the best and underlined the worst average $\Delta Dist$ results.
    }
    \label{tab:fer_results_individual}
\end{table}

\subsubsection{Individual Models}
\label{subsubsec:result_indiv_models_fer}

Similarly to the atttactive attribute, our goal in this sections is to verify whether testing models trained on objective labels of the ExpW dataset actually reduces the fairness metric in the CFD dataset~\cite{chen2021understanding}.
\tabref{tab:fer_results_individual} depicts the average results across five different runs of the individual models, each trained on a different definition of facial expression for the happiness attribute.
We trained one model for each distribution shown in~\figref{fig:data_histograms_fer}.
We also added a row named $H$, which depicts the average result of the models trained on human-based annotations.
As is the case for the attractive attribute, we notice a trade-off between accuracy (`Overall' Accuracy) and fairness ($\Delta Disc$).
This is mainly the case for the models trained using the ObjLCS$_t$ algorithm since it obtains the lowest $\Delta Disc$ values.

However, we see that the best performing ObjLCS algorithm was $t=3$, while the worst was $t=5$ since it has more variability regarding $\Delta Disc$.
Moreover, the algorithm ObjBase obtains competitive accuracy results with the models trained on human labels ($H$), with the expense of having a relatively similar and high fairness measure.
We hypothesize that this happens due to the fact that ObjBase is very strict when it comes to labeling the facial expressions, \ie, it annotates the facial expression only if all the expected AUs, according to literature~\cite{ekman1978action}, are detected.
This hurts the diversity of annotations, as it was described in~\secref{subsubsec:dataset_distribution_fer}.
Additionally, this can be easily visualized in~\figref{fig:hist_objbase}, which labels more than 70\% of the training dataset as containing two facial expressions, \ie, neutral and happiness.
Nonetheless, we can conclude that individually training models on the objective notions of facial expressions might improve the fairness metrics.

\begin{figure*}[!t]
    \centering
    \begin{subfigure}[b]{0.45\textwidth}
        \centering
        \includegraphics[width=1\textwidth,height=4cm]{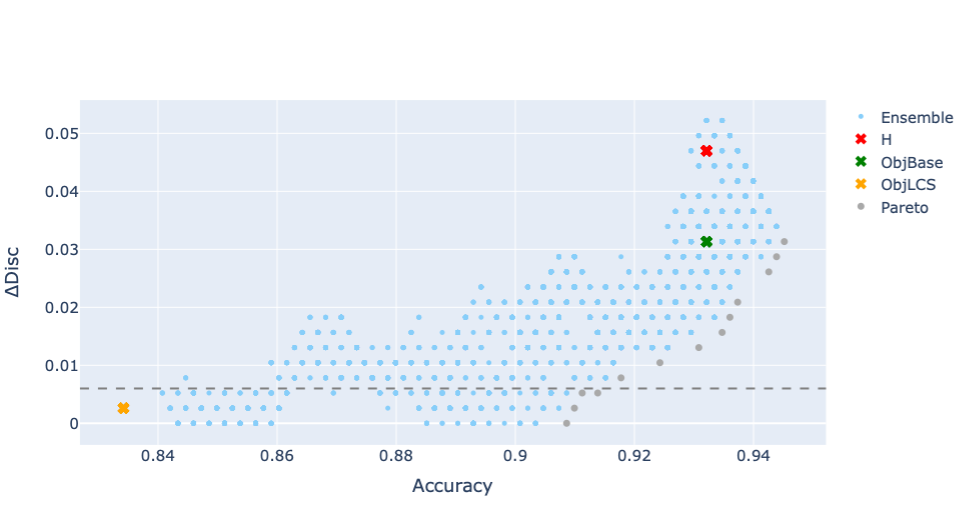}
        \caption{Best performing models.}
        \label{fig:fer_ensemble_best_gender_pareto}
    \end{subfigure}
    ~
    \begin{subfigure}[b]{0.45\textwidth}
        \centering
        \includegraphics[width=1\textwidth,height=4cm]{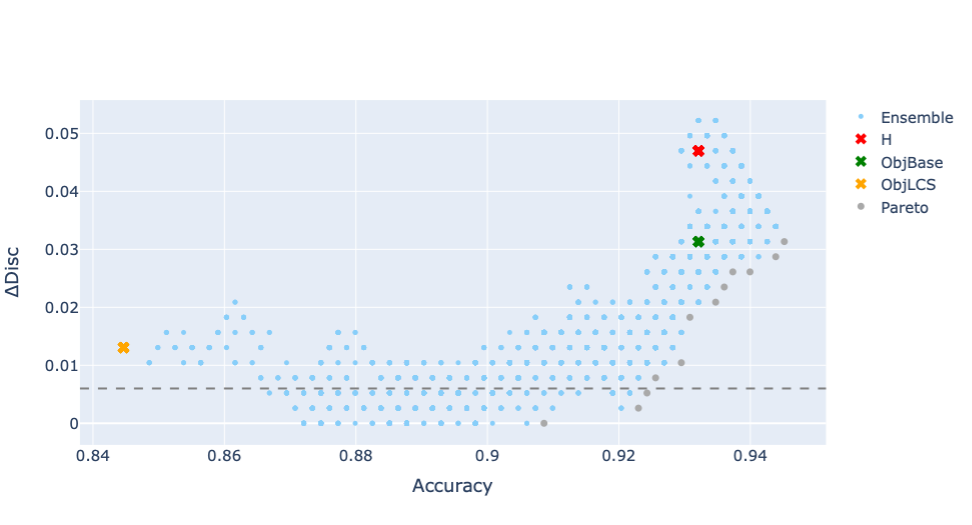}
        \caption{Worst performing models.}
        \label{fig:fer_ensemble_worst_gender_pareto}
    \end{subfigure}
    \caption{
    Results of different weighting values to compose the final ensemble for the facial expression attribute. 
    Plots on the first and second column correspond to the best and worst individual models with respect to $\Delta Dist$ in~\tabref{tab:fer_results_individual}, respectively. 
    We show the result of each ensemble with respect to accuracy ($x$ axis) and $\Delta Dist$ ($y$ axis) for the sensitive attribute \textit{gender}. 
    Light blue dots represent the \textit{ensemble} models, \ie, combining different annotations; red, green, and orange crosses indicate the model trained with only human-based annotations ($H$), base (ObjBase) and LCS (ObjLCS$_t$) objective annotations, respectively; finally, the gray dots represent the Pareto analysis.
    }
    \label{fig:fer_results_ensemble_gender}
\end{figure*}

\subsubsection{Ensemble Model}
\label{subsubsec:result_ensemble_fer}

The ensembles in this section are produced by a weighted combination of three models, \ie, $H$, ObjBase and ObjLCS$_t$, each trained on a different definition of facial expression.
We followed the same procedure as in the attractive attribute, and randomly chose one seed for all models.
\figref{fig:fer_results_ensemble_gender} shows the result of several weighting values per model.
The plot on the left corresponds to the ensemble results for the selected best individual models, \ie, models which obtained best results with respect to $\Delta Dist$,
while the one on the right depicts the result for the worst models.

We show the result of each ensemble with respect to accuracy and $\Delta Disc$ in~\figref{fig:fer_results_ensemble_gender}.
Each blue dot illustrates the result of one ensemble model, and the individual models are shown as crosses: the model trained on human-based annotations is shown in red, and the ones trained only with mathematical concepts, such as ObjBase and ObjLCS$_t$, are shown in green and orange, respectively.
We varied the weight of each individual (base) model from $0$ to $1$ with steps $0.05$.
Finally, the gray dots represent the Pareto analysis.

We observe that both~\figref{fig:fer_ensemble_best_gender_pareto} and~\figref{fig:fer_ensemble_worst_gender_pareto} show comparable curve results, suggesting again that individually combining the best and worst models into an ensemble have approximately the same result regarding overall accuracy and $\Delta Disc$.
We even visualize that the ensembles that combine the worst performing models have more models with $\Delta Disc = 0$.
This is possible due to the fact that this is a less strict metric, relying exclusively in keeping the proportion between predictions of both sensitive groups balanced, \ie, it does not use any information from the labels.
As is the case for the attractive attribute, the final result is heavily dependent on the weights each base model has in the final ensemble.
Thus, from \figref{fig:fer_results_ensemble_gender} we can see that there is a range of possible models, \eg, from the horizontal gray line, fixed at $\Delta Disc=0.006$~\cite{chen2021understanding}, it is possible to obtain an accuracy of $0.9$.

\begin{table}[!t]
    \begin{adjustbox}{width=1\columnwidth}
    \small
    \centering
        \begin{tabular*}{1\columnwidth}{@{\extracolsep{\fill}}l|c|c}
            \toprule
            Def. & Accuracy & $\Delta Disc$ \\
            \midrule
            Baseline~\cite{chen2021understanding} & - & 0.059 $\pm$ 0.035 \\
            Baseline (our $H$) & 0.935 $\pm$ 0.009$^{\ast}$ & 0.046 $\pm$ 0.025$^{\ast}$ \\
            \midrule
            Uniform Confusion~\cite{alvi2018turning} & 0.934 $\pm$ 0.008$^{\ast}$ & 0.046 $\pm$ 0.008 \\
            Gradient Projection~\cite{zhang2018mitigating} & 0.842 $\pm$ 0.107$^{\ast}$ & 0.036 $\pm$ 0.014 \\
            Domain Discriminative~\cite{wang2020towards} & 0.931 $\pm$ 0.013$^{\ast}$ & 0.076 $\pm$ 0.024 \\
            Domain Independent~\cite{wang2020towards} & 0.920 $\pm$ 0.021$^{\ast}$ & 0.029 $\pm$ 0.015 \\
            AUC-FER~\cite{chen2021understanding} & 0.900 $\pm$ 0.009$^{\ast}$ & \underline{0.006} $\pm$ 0.020 \\
            \midrule
            $Ours_1$ & 0.905 $\pm$ 0.008 & \underline{0.006} $\pm$ 0.007 \\ 
            $Ours_2$ & 0.896 $\pm$ 0.010 & \textbf{0.005} $\pm$ 0.006 \\ 
            $Ours_3$ & 0.887 $\pm$ 0.010 & \textbf{0.004} $\pm$ 0.008 \\ 
            $Ours_4$ & 0.862 $\pm$ 0.002 & \textbf{0.001} $\pm$ 0.001 \\ 
            \bottomrule
        \end{tabular*}
    \end{adjustbox}
    \caption{
    Fairness scores for some debiasing approaches.
    We compare against a set of ensembles we obtained from averaging the accuracy and $\Delta Dist$ and running the Pareto analysis on top of it.
    $^{\ast}$ symbol represents the values we obtained when reproducing previous work, according to the code provided by Chen and Joo~\cite{chen2021understanding}.
    }
    \label{tab:fer_comparison_sota}
\end{table}

\subsubsection{Comparison with Prior Work}
\label{subsubsec:comparison_sota_fer}

In this section, we compare our method with previous debiasing approaches for the FER system~\cite{alvi2018turning, zhang2018mitigating, wang2020towards, chen2021understanding}.
Following previous work~\cite{chen2021understanding}, we present results with the average and standard deviation ($\pm$) over all the runs.
Thus, in order to choose some of all possible ensembles, we run the Pareto analysis on the average results, and sort them regarding the fairness metric.
\tabref{tab:fer_comparison_sota} depicts the results.
We note that three of the four selected results have the lowest $\Delta Disc$, and that the one that obtains a similar $\Delta Disc$ as the work of Chen and Joo~\cite{chen2021understanding} also has a competitive accuracy.
Moreover, we show that all of our results obtain a low standard deviation compared to previous methods.
Thus, as is the case for the attractive attribute, we obtained a better trade-off between a given fairness metric ($\Delta Disc$) and accuracy for the FER system.

\section{Ethical Considerations}
\label{sec:ethical_considerations}

The technique proposed in this paper can be applied to mitigate unintended and undesirable biases in some facial analysis systems.
While the idea behind our proposed method is important and can be broadly applied to many other domains, it is not sufficient. 
Rather, as described in Denton et. al~\cite{denton2019image} it must be part of a larger, socially contextualized project to critically assess ethical concerns relating to facial analysis technology.
This project must include addressing questions of whether and when to deploy technologies, frameworks for democratic control and accountability, and design practices which emphasize autonomy, inclusion, and privacy.

Regarding dataset choice, in this work we use CelebA dataset~\cite{liu2015faceattributes} for the attractive attribute, and Expw~\cite{zhang2015learning, zhang2018facial} and CFD~\cite{ma2015chicago} for the facial expression attribute.
All of the attributes within the CelebA dataset are reported as binary categories, and for the ExpW and CFD datasets we follow the procedure on Chen and Joo~\cite{chen2021understanding} to binarize the facial expressions into happy/unhappy.
We note that in many cases this binary categorization does not reflect the real human diversity of attributes.
This is perhaps most notable when the attributes are related to continuous factors.

Moreover, we note that in this work attractiveness and facial expressions were used as means instead of ends.
We do not wish to reinforce any type of prejudice or discrimination based on this measurements, nor motivate inferring these measures for individuals without their consent.
Instead, we use these attributes mainly as applications of our proposed method.
Additionally, gender is not necessarily the one the person identifies with, rather we considered gender expression, which can be often directly inferred by humans.

Finally, we also note that our method may have other limitations.
For instance, we considered datasets collected from in-the-wild images.
These images do not have any background, facial orientation or facial emotion pattern.
Rather, it contains different background colors, frontal and lateral faces, and several facial expressions. 
This may present a limitation, since our method, which is based on landmark and AU extraction, does not fully work on lateral facial poses.


\section{Final Considerations}
\label{sec:final_considerations}

In this paper, we propose a new method for learning fairer models.
Our approach incorporates diversity into the models by combining models trained with subjective (human-based) annotations as well as objective (mathematically-based) labels.
This approach can be extended beyond the tasks explored in this work, and, in general, one can use any objective measures for tasks requiring subjective human labeling within the proposed framework.
Although such objective measures may not always be accurate in practice, the belief is that because these measures are often geometrical attributes
, they are fairer than the subjective labels in the training data and can thus be used 
to mitigate fairness issues.
We demonstrated that our method improves the fairness metrics over the baselines, while maintaining competitive accuracy.
For future work, we intend on analyzing which factors influence the discriminative behavior of the baseline model, as well as expand our work to other sensitive attributes.


\bibliographystyle{IEEEtran}
\bibliography{main}






\end{document}